%% file: AI-identity-report.tex
\documentclass[
	a4paper,
]{VulcanTechReport}

\usepackage{soul}
\usepackage{todonotes}

\newtoggle{showfeedback}
\settoggle{showfeedback}{true}

\iftoggle{showfeedback}{%
  \sethlcolor{vulcanyellow}
}{%
}

\usepackage{amsmath}
\usepackage{multirow}
\usepackage{tikz}
\usetikzlibrary{positioning}
\usetikzlibrary{decorations.pathreplacing}
\usepackage{hyperref}
\usepackage{enumitem}

\colorlet{vulcanyellowpale}{vulcanyellow!20!white}

\newsavebox{\researchboxtmp}
\newenvironment{researchbox}{%
  \par\medskip
  \begin{lrbox}{\researchboxtmp}%
    \begin{minipage}{\dimexpr\linewidth-16pt\relax}%
      \vspace{3pt}%
      \small\textbf{Research directions:}\par
      \begin{itemize}[leftmargin=1.5em, nosep, topsep=4pt, itemsep=2pt, parsep=0pt]%
}{%
      \end{itemize}%
      \vspace{3pt}%
    \end{minipage}%
  \end{lrbox}%
  \noindent{\fboxsep=8pt\relax\colorbox{vulcanyellowpale}{\usebox{\researchboxtmp}}}%
  \par\smallskip
}

\reporttitle{AI Identity: Standards, Gaps, and Research Directions for AI Agents}
\reportsubtitle{Analysis Report}
\newcommand{\affnum}[1]{$^{\text{\robotocondensed #1}}$}
\reportauthors{Takumi Otsuka\affnum{1, 2}, Kentaroh Toyoda\affnum{1, 3}, Alex Leung\affnum{1}}
\newcommand{\reportaffiliations}[1]{\renewcommand{\reportaffiliations}{#1}}
\reportaffiliations{%
  \affnum{1}AIFT, Singapore\\[2pt]%
  \affnum{2}Graduate School of Fundamental Science and Engineering, Department of Computer Science and Communications Engineering, Waseda University, Japan\\[2pt]%
  \affnum{3}Keio Global Research Institute (KGRI), Japan%
}
\reportdate{\today}

\leftheadercontent{}
\rightheadercontent{\includegraphics[width=3cm]{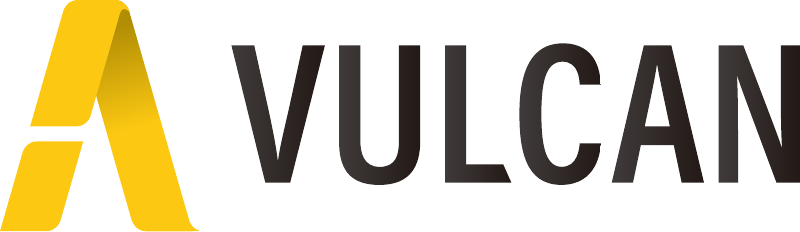}}

\begin{document}

\thispagestyle{empty}
\vspace*{-0.075\textheight}
\hfill\includegraphics[width=5cm]{vulcan-logo.pdf}
\vspace{0.06\textheight}

{\large\raggedright\reportdate\par}
\vspace{0.01\textheight}

{\fontsize{32pt}{34pt}\selectfont\raggedright\textbf{\reporttitle}\par}
\vspace{0.03\textheight}

{\Large\raggedright\textit{\textbf{\reportsubtitle}}\par}
\vspace{0.06\textheight}

{\large\raggedright\reportauthors\par}
\vspace{0.01\textheight}
{\small\raggedright\reportaffiliations\par}
\vspace{0.05\textheight}

\vspace{0.5cm}

\section*{Executive Summary}

\noindent AI agents are now running real transactions, workflows, and sub-agent chains across organizational boundaries, mostly without continuous human supervision. This creates a problem that no current infrastructure is equipped to solve: how do you identify, verify, and hold accountable an entity with no body, no persistent memory, and no legal standing? \textit{AI Identity is the continuous relationship between what an AI agent is declared to be and what it is observed to do, bounded by the confidence that those two things correspond at any given moment.} The identity systems that governed digital life for three decades (passwords, biometrics, and single sign-on) were built for human users, and they have collapsed under the weight of agents that are not human.

To understand what identity infrastructure an agent-saturated world actually requires, we surveyed industry trends, emerging standards, and the technical literature, and conducted a structured gap analysis across the full agent identity lifecycle. This report makes three contributions:
\begin{enumerate}
    \item A structural comparison of human and AI identity across four dimensions: substrate, persistence, verifiability, and legal standing. The asymmetry is fundamental and extending human frameworks to agents without structural modification produces systematic failures (\S\ref{sec:comparison}).
    \item An evaluation of the state-of-the-art technical and regulatory documents against the identity requirements of autonomous agents, revealing that none adequately address the core challenge of governing nondeterministic, boundary-crossing entities (\S\ref{sec:industry}).
    \item Identification of five critical gaps (semantic intent verification, recursive delegation accountability, agent identity integrity, governance opacity and enforcement, and operational sustainability), none of which current technology or regulatory instruments resolve (\S\ref{sec:technologies} to \ref{sec:gaps}).
\end{enumerate}

These gaps are structural; more engineering effort alone will not close them. Foundational research on AI identity is therefore the central conclusion of this report.

\section{Introduction}\label{sec:intro}

\noindent AI agents have moved from research demonstrations to operational infrastructure. Enterprises across finance, healthcare, and enterprise IT now run autonomous agents that chain multi-step API (Application Programming Interface) calls via protocols such as MCP (Model Context Protocol) and A2A (Agent2Agent), spawn sub-agents, and act across organizational boundaries without continuous human supervision \cite{gravitee2026, authworkflows2026}. Unlike the scripts and bots of earlier generations, modern agents acquire credentials, accumulate context through memory systems, and take irreversible actions at machine speed. This shift is outpacing the governance frameworks meant to oversee it.

That infrastructure has collapsed under three concurrent failures.

\textbf{Organization}: non-human identities now outnumber human identities in enterprise environments (a recent analysis of 27 million enterprise NHIs (Non-Human Identities) found a ratio of 144 to 1~\cite{entro2025}), yet only 21.9\% of organizations treat agents as independent identity principals; the remainder run agents on shared API keys (45.6\%) or inherited human credentials never designed for non-human use~\cite{gravitee2026}; OAuth and SAML (Security Assertion Markup Language), designed for humans at browsers, are structurally unable to govern agents that spawn sub-agents without human-in-the-loop oversight.

\textbf{Regulation}: no jurisdiction has established a liability framework for autonomous AI agents: the EU (European Union) AI Act classifies risk but does not assign responsibility when a delegated agent acts outside its mandate~\cite{euaiact50}; the US (United States) has no federal AI identity legislation; no technical standard covers the full delegation lifecycle.

\textbf{Technology}: agents are nondeterministic: the same model weights produce different outputs even on the same inputs, so a credential that verifies what an agent \textit{is} cannot guarantee what it will \textit{do}; agents are cloneable: model weights can be copied and run across many concurrent instances, making instance uniqueness unenforceable without hardware binding; and agents can be sessionless: lacking persistent memory across interactions by default, there is no stable substrate to anchor a continuous identity to. 

\noindent These concurrent failures define the central question this report addresses: \textit{what identity infrastructure does an agent-saturated world require?} We address it through four research sub-questions, each answered in a dedicated section:

\begin{enumerate}[label=\textbf{RQ\arabic*.}]
    \item How should identity for AI agents fundamentally differ from identity for humans, and what should be the structural consequences of that asymmetry? (\S\ref{sec:comparison})
    \item Do current market solutions and emerging standards adequately address the identity requirements of autonomous, nondeterministic AI agents? (\S\ref{sec:industry})
    \item What technologies are available to address the identity requirements of autonomous AI agents, and where do current solutions fall short? (\S\ref{sec:technologies})
    \item What structural gaps persist across all current approaches, what are the hard boundaries of existing solutions, and what research directions does each gap motivate? (\S\ref{sec:gaps})
\end{enumerate}

Answering these questions directly benefits both industry and research: practitioners gain structured criteria for evaluating and deploying agent identity infrastructure, standards bodies gain a precise map of where current proposals fall short, and researchers gain a set of hard open problems with clear boundaries, the starting conditions for rigorous scientific progress.

To scope the analysis, we searched academic literature, standards corpora (e.g., IETF, OpenID Foundation, W3C, and NIST), regulatory documents, and gray literature including industry reports, market analyses, and vendor whitepapers. Search terms combined identity-related concepts (such as non-human identity, authentication, authorization, delegation, and workload identity) with agent-related terms (such as AI agent, agentic AI, autonomous agent, and multi-agent system). Sources were prioritized from 2024--2026 to capture the rapidly evolving landscape, supplemented by foundational prior work where necessary. In total, approximately 80 sources were reviewed; those directly supporting the analysis are cited throughout the report.

The remainder of the report proceeds as follows. Section~\ref{sec:comparison} establishes the conceptual foundation by comparing human and AI identity across four structural dimensions. Section~\ref{sec:industry} maps the market landscape, standards trajectory, and regulatory context. Section~\ref{sec:technologies} surveys the technologies that the market has proposed as solutions and identifies critical gaps where those solutions fail. Section~\ref{sec:gaps} analyzes the structural nature of those gaps and embeds research directions within each gap subsection. Section~\ref{sec:conclusion} synthesizes the findings and proposes a unifying theoretical frame for the research agenda.

\section{RQ1: Comparison of Human and Non-Human Identities}\label{sec:comparison}

\noindent Before surveying technologies or markets, we need to understand what identity means for humans and machines, and why the difference is structural rather than incremental. This section builds the conceptual foundation that everything else depends on.

\subsection{Human Identity}\label{subsec:human}

Human identity rests on a biological substrate, namely DNA, neural tissue, and the physiological continuity of a living body, that persists across time and context \cite{aipersonhoodtheory2025}.
A person remains recognizably the same individual through sleep, conversation, and shifts in social role. This continuity anchors every major identity system. Biometric verification assumes that fingerprints, facial geometry, and iris patterns are stable signatures of a unique biological entity. Social institutions, including birth certificates, passports, and employment records, assume a persistent subject to whom rights and obligations attach. Legal frameworks reinforce this by treating identity as both a fundamental right (e.g., GDPR (General Data Protection Regulation) \cite{gdpr2016}, eIDAS~2.0 (Electronic Identification, Authentication and Trust Services) \cite{eidas2024}) and, increasingly, as commercial property subject to explicit consent and control \cite{ab2602, shtefan2025}.

\subsection{Non-Human Identity}\label{subsec:nhi}

\begin{table}[ht]
\centering
\caption{Summary of non-human identity types.}
\label{tab:nhi-types}
\small
\begin{tabular}{p{1.8cm} p{5.0cm} p{4.2cm} p{2.8cm}}
\hline
\textbf{Type} & \textbf{What it represents} & \textbf{Typical artifact} & \textbf{Lifecycle} \\
\hline
Model & Trained artifact: weights, architecture, provenance & Model card, weight hash & Stable until retrained \\[4pt]
Agent & Configured deployment: system prompt, persona, tool grants & Agent card, system prompt hash & Session\,/\,task \\[4pt]
Workload & Running process\,/\,container\,/\,service instance & SPIFFE SVID (X.509\,/\,JWT) & Ephemeral \\[4pt]
Delegated & Authorization derived from a human\,/\,org principal & OAuth token, JWT \texttt{act} claim & Scoped to grant \\[4pt]
\hline
\end{tabular}
\end{table}

This biological foundation, and the continuity and legal standing that rest on it, has no counterpart in AI. There is no biological substrate, no subjective continuity, and no inherent social standing. It is useful to separate NHI into four distinct types, because conflating them leads to mismatched solutions (Table~\ref{tab:nhi-types}).

At the \textit{model level}, identity is the trained artifact itself: the \textit{weights} (learned behavior), the \textit{architecture} (how those weights process inputs), and the training provenance (what data shaped them). These properties stay stable between deployments, but they are not uniquely identifying in the way a fingerprint is: two deployments of the same weights are the same model-level identity, even if they behave differently in context \cite{personhoodcredentials2024, aipersonhoodtheory2025, pragmaticpersonhood2025}. Model cards \cite{modelcards2019} are the closest existing artifact to a model-level identity document, recording training data, intended use, and evaluation results. They are, however, descriptive and non-binding: a model card is a disclosure, not a verifiable credential, and it provides no cryptographic guarantee that a deployed model corresponds to what the card describes.

At the \textit{agent level}, identity is constituted by the configuration and runtime state that shapes a specific deployment: a \textit{system prompt} that defines the agent's role and constraints but is manipulable via prompt injection; a \textit{persona or behavioral specification} (such as OpenClaw's \texttt{SOUL.md} file) that encodes persistent values and goals; the \textit{tool calls and capabilities} the agent has been granted access to; and externally assigned \textit{credentials} that can be revoked, rotated, or shared \cite{authworkflows2026}. All of these are mutable, and none is uniquely tied to a singular entity in the way a fingerprint is tied to a body. This level is where most identity-related vulnerabilities arise: a system prompt can be injected, credentials can be stolen, and tool access can be abused without any change to the underlying model.

At the \textit{workload level}, identity is the runtime instance of a process, container, or service: not which model or which configuration, but \textit{which specific execution} is on the wire right now, on which host, under which cloud account. The operative credential is an ephemeral, attestation-bound certificate or JWT (JSON Web Token) issued by a workload identity authority such as SPIFFE (Secure Production Identity Framework For Everyone)/SPIRE (SPIFFE Runtime Environment), which verifies properties of the requesting process before issuing it an SVID (SPIFFE Verifiable Identity Document) \cite{ietfdraft2026}. The same agent configuration may run as thousands of simultaneous workload instances; workload identity distinguishes them at the infrastructure layer without requiring per-instance enrollment. Its critical limitation is scope: it authenticates the container and not the content, proving that a trusted process is on the wire but providing no guarantee about what that process will do next.

At the \textit{delegated level}, identity is constituted not by what an entity \textit{is} but by what it has been \textit{authorized to do on behalf of} a human or organizational principal. A delegated identity is carried by a verifiable grant, such as an OAuth~2.0 access token, a JWT (JSON Web Token) with an \texttt{act} (actor) claim, or a signed capability document, that traces the authorization chain back to a principal with legal standing \cite{authworkflows2026, openid2025}. The defining constraint is \textit{scope attenuation}: each delegation hop must narrow, never widen, the set of permitted actions, so that no sub-agent can accumulate capabilities the original human principal did not authorize. Delegated identity is not a peer type in the same sense as the other three; any model, agent, or workload identity can carry a delegated grant. It is named separately here because the \textit{absence} of a deployed standard for multi-hop delegation chain verification is the central accountability gap examined in \S\ref{subsec:authz} and \S\ref{sec:gaps}.

\subsection{A Four-Dimension Comparison}\label{subsec:comparison-table}

\begin{table}[ht]
\centering
\caption{Four-dimension comparison of human and AI identity.}
\label{tab:identity-comparison}
\small
\begin{tabular}{p{2.4cm} p{4.8cm} p{5.8cm}}
\hline
\textbf{Dimension} & \textbf{Human} & \textbf{AI Agent} \\
\hline
Substrate & Biological (DNA, neural tissue) & Computational (model weights, architecture, system prompt, credentials; all mutable) \\[4pt]
Persistence & Continuous across time and context & Continuous at the model level, but operational context resets frequently; shorter lifecycle and higher sensitivity to configuration changes than human identity \\[4pt]
Verifiability & Biometric features serve as stable anchors for point-in-time verification & No stable behavioral anchor; point-in-time verification is inherently incomplete because identical inputs can produce different outputs (due to temperature sampling and context drift), and configuration changes alter behavior between enrollment and action \\[4pt]
Legal standing & Fundamental right (GDPR) and commercial property (AB~2602, ELVIS Act) & No legal personhood; identity is delegated from a human or organizational principal \\[4pt]
\hline
\end{tabular}
\end{table}

To make the structural asymmetry precise, Table~\ref{tab:identity-comparison} compares human and AI identity across four key dimensions. \textit{It is clear that the identity infrastructure required for AI agents cannot be derived by relaxing or extending the infrastructure designed for humans.} It must be designed from first principles, with the four asymmetries in Table~\ref{tab:identity-comparison} treated as architectural constraints rather than edge cases. How the market and standards community have begun to respond to this challenge is the subject of the next section.

\section{RQ2: Industry Trends}\label{sec:industry}

\noindent With the conceptual distinction between human and AI identity established, this section surveys the commercial, standards, and regulatory context in which AI identity is developing. We look at how vendors are positioning, what standards bodies have proposed, and what regulators are requiring, to assess whether the response to the structural asymmetry identified in \S\ref{sec:comparison} is adequate.

\subsection{Vendor Direction and Emerging Players}\label{subsec:vendors}

The vendor landscape can be read as a set of partial answers to distinct stages of the agent identity lifecycle, with no single product spanning enrollment, runtime authorization, and behavioral accountability.

\textbf{Governance and lifecycle consolidation.} One cluster addresses the enrollment and lifecycle problem: who provisions an agent, under what policy, and how is it decommissioned. Saviynt \cite{saviynt2025} unifies human IAM (Identity and Access Management), machine identity, and privileged access under a single governance platform, while RadiantOne \cite{radiantone2025} provides correlation across identity silos so that agent accounts do not fragment into shadow inventories. Astrix has been recognized by Gartner for NHI-specific posture management \cite{astrix2025}, focusing on discovery and risk scoring of existing non-human principals. Coordinating this direction, the NHIMG (Non-Human Identity Management Group) \cite{nhimg2025} is defining governance baselines that treat agents as first-class principals rather than elevated service accounts. These tools solve visibility and policy attachment, but operate above the credential layer and assume the existence of a trustworthy identity to govern.

\textbf{Runtime credentials and trust verification.} A second cluster targets the issuance and verification problem: producing cryptographic identities an agent can present at runtime and evaluating whether a counterparty should accept them. HashiCorp's Vault~1.21 introduced native SPIFFE (Secure Production Identity Framework For Everyone) authentication \cite{hashicorp2025}, enabling ephemeral, attestation-bound workload identities issued directly within the secrets management layer. Vouched's KYA (Know Your Agent) platform \cite{vouched2025a} extends KYC (Know Your Customer)-style verification to autonomous software. HUMAN Security's AgenticTrust \cite{humansecurity2025} sits at the relying-party side, treating the boundary between a malicious bot and an authorized agent as a contextual judgment over provenance and delegation rather than a network-signature problem. Collectively, these products issue, verify, and govern agent identities, but they do so against incompatible credential formats, trust roots, and delegation semantics, a fragmentation that motivates the standards survey in the next section.

\subsection{Standards Landscape}\label{subsec:standards}

\begin{table}[ht]
\centering
\small
\caption{Standards evaluation for agentic identity, grouped by document type.}
\label{tab:standards}
\begin{tabular}{lp{3.2cm}lllp{2.6cm}}
\toprule
\textbf{Category} & \textbf{Standard} & \textbf{Verdict} & \textbf{Key Limitation} \\
\midrule
\multirow{5}{*}{\textbf{Protocol}}
                  & OAuth 2.0         & Partial     & No multi-hop or scope-to-skill mapping \\
                  & SAML              & Fails       & Session-based; assumes human browser \\
                  & WIMSE/SPIFFE      & Available   & Identity layer only; no authorization \\
                  & MCP               & Partial     & Identity out of scope by design \\
                  & A2A               & Partial     & JWS integrity only; token lifetime, scope, and consent gaps \\
\midrule
\multirow{2}{*}{\textbf{Specification}}
                  & OpenID Agentic AI & Partial     & CIBA cannot scale; two-tiered web risk \\
                  & IETF AIMS         & Partial     & Architecture draft composing existing standards; security sections TODO \\
\midrule
\textbf{Regulation}
                  & eIDAS 2.0         & Infra/Human & Wallet mandate; designed for citizens \\
\midrule
\multirow{3}{*}{\textbf{Guideline}}
                  & NIST NCCoE        & Directional & Concept paper; no implementation guidance \\
                  & OWASP Top~10 for Agentic Applications  & Directional  & Top~10 risks with mitigation guidelines; no identity protocol \\
\bottomrule
\end{tabular}
\end{table}

\noindent We evaluated technical and regulatory documents by asking what function each performs in the AI identity lifecycle (Table~\ref{tab:standards}): How does it handle \textit{authentication}, \textit{authorization and delegation}, \textit{credential issuance and portability}, and \textit{governance and diagnostics}? For each, we assigned a verdict: \textit{Available} (production-ready), \textit{Partial} (structural gaps), \textit{Fails} (architecturally incompatible), \textit{Directional} (conceptually sound but no implementation guidance), or \textit{Diagnostic} (identifies problems without solutions). Table~\ref{tab:standards} summarizes the results. The structural pattern is clear: no single standard spans the full agent identity lifecycle, and the gaps are not random: they cluster at the hardest parts of the problem.

For \textbf{authentication}, WIMSE (Workload Identity in Multi-cloud and Service Environments)/SPIFFE is the most mature option, providing ephemeral, attestation-bound workload credentials at scale \cite{ietfdraft2026} (see~\S\ref{subsec:auth}). The IETF AIMS draft (an individual submission to the WIMSE working group) offers a more comprehensive vision including dual-identity credentials and explicit agent lifecycle management, but leaves key security considerations unresolved. SAML fails outright: its session-based, browser-mediated architecture cannot accommodate non-human principals.

For \textbf{authorization and delegation}, OAuth~2.0 handles one-hop delegation well but lacks multi-hop chaining, cross-domain asynchronous flows, and any mapping between OAuth scopes and agent capabilities. MCP has achieved remarkable adoption \cite{openid2025} and added OAuth~2.1 support, but it suffers from specific authorization failures (see~\S\ref{subsec:authz}). Google's A2A provides agent discovery and integrity via JWS (JSON Web Signature) signing but delegates all authorization and identity decisions to other protocols. The OpenID Foundation's proposed OBO (On-Behalf-Of) flows acknowledge the impersonation accountability gap, but reliance on CIBA (Client-Initiated Backchannel Authentication) for human-in-the-loop consent cannot scale to autonomous agent throughput.

For \textbf{governance and diagnostics}, eIDAS~2.0 mandates EU-wide wallet infrastructure but targets citizens rather than agents. The NIST (National Institute of Standards and Technology) NCCoE (National Cybersecurity Center of Excellence) has published a concept paper identifying five focus areas but no implementation guidance \cite{nist2026}. OWASP's (Open Web Application Security Project) Top~10 for Agentic Applications catalogs the ten highest-impact agentic security risks and provides mitigation guidelines for each, but does not define identity protocols or prescribe implementable authorization controls \cite{owasptop102025}. These three are valuable as orientation documents, but the absence of implementation guidance means practitioners cannot derive concrete controls from them.

The overall picture that emerges is that authentication is the most mature function, authorization and delegation are partially addressed for simple cases and completely unaddressed for multi-hop chains, and governance remains largely diagnostic.

\subsection{Regulatory Landscape}\label{subsec:regulatory}

\begin{table}[ht]
\centering
\small
\caption{AI identity regulation across five jurisdictions.}
\label{tab:regulatory}
\begin{tabular}{@{}lp{12cm}@{}}
\toprule
\textbf{Jurisdiction} & \textbf{Key Instruments} \\
\midrule
EU         & AI Act Art.~50 (transparency obligations, effective Aug 2026)~\cite{euaiact50}; Code of Practice on AI-generated content (Dec 2025)~\cite{euaioffice_cop2025}; eIDAS~2.0 EUDI Wallet~\cite{eidas2024}; Cyber Resilience Act (Dec 2024)~\cite{eucra2024} \\
US         & NIST NCCoE concept paper~\cite{nist2026} (5 focus areas); CAISI AI Agent Standards Initiative (Feb 2026)~\cite{nistcaisi2025} (3 pillars); OMB M-25-21~\cite{omb_m2521_2025} (high-impact AI categories); no federal AI identity law \\
China      & Measures for Labeling AI-Generated Content (Sep 2025, GB~45438-2025)~\cite{china_ailabeling2025}; CAC Generative AI Measures (Aug 2023)~\cite{cac_genai2023}; Revised Cybersecurity Law Art.~20~\cite{chinacybersec2026}; draft virtual-human rules (authentication-bypass prohibition) \\
Japan      & AI Promotion Act (May 2025, innovation-first, no binding obligations)~\cite{japan_aiact2025}; METI/MIC AI Business Guidelines v1.1 (Mar 2025)~\cite{meti_aiguide2025}; Hiroshima AI Process Reporting Framework (Feb 2025)~\cite{hiroshima_aip2025} \\
Singapore  & IMDA Model Governance Framework for Agentic AI (Jan 2026) \cite{imda2026}; four governance dimensions. CSA Addendum on Securing Agentic AI (Oct 2025) \cite{csa_addendum2025}; threat T9 (identity spoofing) and authentication controls \\
\bottomrule
\end{tabular}
\end{table}

\noindent We studied the regulatory landscape across five major jurisdictions to understand what legal obligations an AI agent operating internationally must satisfy and where the gaps are. The most consequential structural challenge for AI identity is not that regulation is absent but that it is fragmented: an agent that complies with the EU's transparency requirements under AI Act Article~50 \cite{euaiact50} may violate China's mandatory content labeling rules. Harmonization will not resolve this: the jurisdictions disagree at a structural level about what problem AI identity regulation is meant to solve (Table~\ref{tab:regulatory}).

The \textbf{EU} AI Act Article~50 (effective August 2026) requires machine-readable marking of AI-generated content and real-time disclosure to users~\cite{euaiact50}. The EU AI Office is developing a Code of Practice to implement these obligations~\cite{euaioffice_cop2025}. Two instruments address the underlying identity infrastructure: eIDAS~2.0 (Regulation 2024/1183) establishes the EUDI (European Digital Identity) Wallet whose cryptographic foundations could extend to AI-agent authorization through government-issued digital identities~\cite{eidas2024}, and the CRA (Cyber Resilience Act, in force December 2024) mandates cybersecurity controls for products with digital elements, with AI systems satisfying CRA requirements deemed compliant with AI Act Article~15~\cite{eucra2024}.

The \textbf{US} has no federal AI identity legislation. The NIST NCCoE concept paper (February 2026) directly targets the gap, identifying agent authentication, zero-trust authorization, non-repudiation, prompt injection controls, and governance as the five focus areas for AI agent identity management~\cite{nist2026}; the CAISI (Center for AI Standards and Innovation) AI Agent Standards Initiative (launched February 2026) is translating these into interoperability standards~\cite{nistcaisi2025}. OMB Memorandum M-25-21 establishes pre-deployment testing and human oversight requirements for federal agencies deploying biometric AI identification systems, but carries no private-sector mandate~\cite{omb_m2521_2025}.

\textbf{China} addresses AI identity through output attribution and service registration. The CAC (Cyberspace Administration of China) Measures for Labeling AI-Generated Content (effective September 2025) mandate machine-readable metadata that attributes AI-generated output to the producing system, alongside visible indicators, backed by mandatory national standard GB~45438-2025~\cite{china_ailabeling2025}; the CAC Generative AI filing regime creates an official identity record for each registered public-facing AI service~\cite{cac_genai2023}. Draft measures on AI virtual-human services further prohibit synthetic personas from bypassing biometric authentication mechanisms, targeting AI-enabled identity fraud directly.

\textbf{Japan} takes a principles-based, non-binding approach. The AI Promotion Act (enacted May 2025) enshrines transparency as a statutory principle but imposes no obligations or penalties~\cite{japan_aiact2025}, and voluntary METI/MIC AI Business Guidelines v1.1 (March 2025) operationalize governance expectations covering security and accountability for AI providers~\cite{meti_aiguide2025}. Japan also leads the G7 Hiroshima AI Process (HAIP) Reporting Framework (February 2025), through which participating developers publicly disclose their AI practices~\cite{hiroshima_aip2025}, and operates Gennai (``Government AI,'' May 2025), a secure AI environment for approximately 180{,}000 civil servants built to ISMAP (Information Security Management and Assessment Program) security standards~\cite{japandigital2025}.

\textbf{Singapore} deploys two complementary instruments. The IMDA (Infocomm Media Development Authority) Model Governance Framework for Agentic AI, launched at WEF (World Economic Forum) Davos in January 2026, addresses four governance dimensions (accountability, transparency, human oversight, and data governance), providing principles-based guidance for agentic AI deployments~\cite{imda2026}. The CSA (Cyber Security Agency of Singapore) Addendum on Securing Agentic AI (October 2025) takes a cybersecurity angle directly relevant to AI identity: it designates identity spoofing and impersonation as threat T9, requires organizations to maintain a trusted agent registry and authenticate agents using verifiable credentials with short-lived OAuth~2.0/OIDC tokens, and prohibits cross-agent privilege delegation unless explicitly authorized~\cite{csa_addendum2025}. A companion discussion paper identifies agent identity and delegation schemes as an architecturally unresolved gap and calls for standardized identity protocols as a research priority~\cite{csa_discussion2025}.

\section{RQ3: Technologies for AI Identity}\label{sec:technologies}

\noindent Despite this market momentum, rapid standardization activity, and regulatory attention across jurisdictions, the technologies that underpin AI identity have structural limitations that market investment alone cannot resolve. This section surveys the technologies available for each function of agent identity management, organized around six capabilities: authentication, authorization and delegation, credentials and portable identity, provenance and content integrity, governance and monitoring, and audit logging and attestation, tracing how each addresses the identity requirements identified in \S\ref{sec:comparison}.

\subsection{Authentication}\label{subsec:auth}

Authenticating an AI agent differs structurally from authenticating a human user. Human authentication ultimately rests on something the person \textit{is} (a biometric), \textit{has} (a device), or \textit{knows} (a secret), anchored to a single embodied identity that persists across sessions and resists cloning. AI agents satisfy none of these properties. An agent instance is \textit{nondeterministic}: the same prompt may yield materially different actions on successive invocations, so behavior cannot be used as an identity signal. It is \textit{cloneable}: a container image, a model checkpoint, or an API key can be replicated indefinitely with no cryptographic distinction between the copies. It is frequently \textit{sessionless}: short-lived function invocations or one-shot tool calls may begin and end faster than conventional session lifetimes. And it has no \textit{biometric anchor} of any kind. These properties force authentication to shift from verifying a persistent embodied subject to attesting ephemeral, replaceable workloads and binding them cryptographically to the humans or organizations on whose behalf they act \cite{openid2025, nhimg2025, zerotrust2025}.

The dominant response in the standards community has been to treat agents as \textit{workloads} rather than users. The IETF (Internet Engineering Task Force) AIMS draft (\texttt{draft-klrc-aiagent-auth-00}), an individual submission to the WIMSE working group, formalizes this posture and composes two layers of identity primitive \cite{ietfdraft2026}. At the transport layer, agents are issued \textit{SVIDs} (SPIFFE Verifiable Identity Documents, either X.509 certificates or JWTs (JSON Web Tokens)) whose subject is a SPIFFE ID of the form \texttt{spiffe://trust-domain/path}. These SVIDs are provisioned by SPIRE (SPIFFE Runtime Environment), the reference implementation of SPIFFE, which runs an agent daemon on each node and performs \textit{workload attestation}: before issuing an SVID, SPIRE verifies properties of the requesting process (its Unix UID, its Kubernetes service account, its container image digest, its cloud instance metadata) against selectors registered for that identity, so a rogue process on the same host cannot simply ask for another workload's credential. Two SPIFFE-identified peers then establish mutual TLS (Transport Layer Security) directly, without a shared secret or a human-mediated enrollment step. Block has deployed the full SPIFFE+WIMSE+OAuth stack in production, offering one of the first real-world validations that the standards-based workload-identity approach is operationally viable \cite{blockwimse2025}. HashiCorp Vault~1.21 added native SPIFFE authentication, allowing an agent that already holds an SVID to exchange it directly for a Vault token and retrieve secrets without a separately managed credential, closing one of the classic bootstrap gaps for AI agents that must reach into secret stores during a task \cite{hashicorp2025}. The NHIMG (Non-Human Identity Management Group) complements this plumbing with enrollment, lifecycle, and revocation practices specific to non-human identities, arguing that workload-style credentials must be accompanied by an authoritative registry and an owner of record for every issued identity \cite{nhimg2025}.

Mutual TLS alone, however, is insufficient for agent authentication. It authenticates the \textit{channel} between two endpoints, but an agent call often traverses multiple hops (an orchestrator, a tool gateway, a downstream API) and the original caller's identity is invisible to anything past the first TLS terminator. To carry identity end-to-end at the application layer, AIMS layers \textit{WIMSE Proof Tokens} over the transport channel \cite{ietfdraft2026}. Unlike a bearer OAuth access token, which any party that captures it can replay, a proof token is \textit{proof-of-possession}: it is cryptographically bound to a key held by the legitimate agent, and each use requires a fresh signature over request-specific data (method, URI, timestamp, nonce), so a stolen token is unusable without the corresponding private key. AIMS further introduces \textit{dual-identity credentials} that bind the agent to its human or organizational owner through three delegation flows (Agent-Mediate, Owner-Mediate, and Server-Mediate), each producing an auditable chain of accountability back to a human principal \cite{ietfdraft2026, auth0video2026}.

For agent-to-agent discovery and authentication across organizational boundaries, Google's A2A protocol takes a lighter-weight, Web-native approach. An agent publishes an \textit{Agent Card}, a JSON document served at \texttt{/.well-known/agent.json}, which advertises its name, endpoint, supported skills, and authentication requirements \cite{googlea2a2025}. The card may be signed using JWS, giving a remote caller integrity over the advertised metadata and a cryptographic link to a publisher key. MCP, which standardizes how agents connect to tools and data sources, added OAuth~2.1 support for its HTTP transport in January~2026, adopting a tightened profile that mandates PKCE (Proof Key for Code Exchange), prohibits the implicit and password grants, and imposes stricter redirect handling \cite{openid2025}. Together, A2A and MCP with OAuth~2.1 cover the two most common agent interaction surfaces: agents talking to other agents, and agents talking to tools. Their limitations, however, are as significant as their contributions. Agent Cards are \textit{self-declared}: the document asserts what an agent can do and who operates it, but no third party attests to model provenance, training lineage, or behavioral conformance, so a malicious or misconfigured operator can publish any card it chooses \cite{gravitee2026}. OAuth~2.1 terminates at the resource server boundary: it confirms that a token is valid and that the agent holds the right scopes, but it does not prevent \textit{confused-deputy} patterns in which a correctly authenticated agent is induced by attacker-controlled input to exercise its privileges against the wrong target \cite{owaspthreats2025, authworkflows2026}. And the IETF AIMS specification's own Security Considerations section still reads ``TODO Security'' in the current draft, signaling that even the authors regard the threat model as unfinished \cite{ietfdraft2026}.

Despite rapid enterprise adoption of non-human identity tooling \cite{saviynt2025, gravitee2026}, a structural gap persists across all of these mechanisms. Every current mechanism authenticates the \textit{container} of identity (a token, a certificate, an SVID, a signed agent card) and not the \textit{content} of the agent: the model weights that produce its behavior, the system prompt that constrains its mandate, or the intent behind its next tool call. A perfectly authenticated agent, holding a valid SVID and a fresh proof-of-possession token, can still act outside its mandate the moment after authentication completes, whether through a prompt injection that rewrites its goals, a hallucinated tool invocation, or a drift in model behavior between versions. Authentication, as currently standardized, establishes \textit{who} is on the wire; it does not establish \textit{what} that party is about to do, and it provides no primitive by which a relying party can decide whether the next action falls inside or outside the agent's authorized scope.

\subsection{Authorization and Delegation}\label{subsec:authz}

Authentication establishes who an entity is; authorization determines what it may do. The dominant human-centric authorization frameworks (OAuth~2.0, OpenID Connect, and SAML) are not merely \textit{designed for humans} in a loose sense: they encode a small number of very specific structural assumptions that autonomous agents violate by construction. OAuth~2.0 presumes a \textit{synchronous human consent} event and \textit{single-hop delegation} from one client to one resource server; SAML's assertion model assumes a bounded interactive \textit{session} anchored to a browser cookie \cite{openid2025, strata2025}. Agents violate all of these: they act asynchronously, often long after any human interaction, and chain calls across multiple services in a single task. The result is that applying OAuth~2.0 to agents without modification either forces impersonation (in which the agent simply replays the user's token) or overly broad static API keys, both of which destroy the accountability properties that made OAuth valuable in the first place \cite{openid2025, zerotrust2025, gravitee2026}.

The OpenID Foundation's \textit{OBO} (On-Behalf-Of) flows address the first of these gaps by replacing direct impersonation with a token-exchange protocol \cite{openid2025, strata2025}. Technically, an OBO token is a new access token minted by the authorization server in response to a token-exchange request in which the calling agent presents both its own client credential and the user's original token; the resulting token carries the original user's identity and the delegating agent's identity as \textit{separate} claims, typically expressed as distinct \texttt{sub} and \texttt{act} (actor) fields, so a downstream resource server can verify the full principal chain \cite{openid2025, strata2025}. \textit{CIBA}, also standardized by the OpenID Foundation \cite{auth0video2026, openid2025}, addresses the synchrony gap by decoupling the device that initiates an authorization request from the device on which the user approves it. In a CIBA flow, the agent sends an authorization request directly to the identity provider (IdP) over a back channel; the IdP then pushes a consent prompt to the user's pre-registered device; the user approves out-of-band; and the IdP subsequently issues a token to the agent. This matters for agents specifically because their actions are often initiated long after the user last interacted with any session, and because a human-in-the-loop approval step must be reachable without assuming that a browser session is still live \cite{auth0video2026}.

Two further architectural patterns are emerging around these flows. The \textit{Triangle of Trust} model \cite{auth0video2026} formalizes the three-way relationship between user, agent, and service as three bilateral trust relationships that must all be independently established: user-agent, agent-service, and user-service. Each pair authenticates on its own footing, so that no single link can be bypassed by leveraging another; an agent cannot act on a service using only the user's trust relationship with that service, because the service must itself have independently verified the agent. \textit{Token Vault} architectures \cite{hashicorp2025, auth0video2026} attack the credential-handling problem from a different angle. Rather than issuing tokens directly to agents, a vault holds the underlying credentials and exposes only opaque references, or handles, to callers. When an agent needs to invoke an API, it presents a handle to the vault, which either resolves the handle to a credential and makes the outbound call itself or returns a short-lived derived token scoped to that single invocation. Because the agent never holds the raw credential, exfiltration yields nothing usable and revocation reduces to invalidating a handle \cite{hashicorp2025}.

The harder problem, and the one that these mechanisms only partially address, is \textit{multi-hop delegation}: when Agent~A delegates to Agent~B, which in turn delegates to Agent~C, no production-ready standard traces the authorization chain back to the originating human principal in a way that every resource server along the chain can verify. The governing principle, \textit{scope attenuation}, holds that each delegation step must narrow, never widen, the set of permitted actions, so that capabilities monotonically shrink down the chain \cite{authworkflows2026, openid2025}. Enforcing scope attenuation in production is difficult for two reasons. First, there is no widely deployed token format that carries an explicit, cryptographically verifiable delegation chain: the IETF AIMS draft \cite{ietfdraft2026} is still at an early stage and no production-ready standard has yet emerged. Second, even where a chain can be represented, resource servers lack a shared vocabulary for comparing scope claims across hops, so an attenuation check that should reject a widening step cannot be mechanically performed. OWASP's Top~10 for Agentic Applications \cite{owasptop102025} flags identity and privilege abuse (ASI03), which explicitly covers exploitation of delegation chains and role inheritance, among its headline risks, and both NIST's NCCoE project on software and AI identity \cite{nist2026} and industry baselines for non-human identity governance \cite{nhimg2025} flag the absence of a standard delegation-chain format as a blocking gap for enterprise adoption.

The Model Context Protocol compounds the difficulty, and the literature on agentic threats has crystallized a specific catalog of its authorization failure modes \cite{owasptop102025}. First, there is \textit{no per-tool authentication}: once an agent authenticates to an MCP server, it implicitly gains access to every tool that server exposes, with no per-tool credential check. Second, \textit{confused-deputy risk} (\S\ref{subsec:auth}) arises because tools share the server's execution context, so a tool invoked with low-sensitivity input can be induced to perform high-sensitivity actions using the server's ambient privileges. Third, \textit{privilege concentration} places all tool capabilities behind a single MCP server whose compromise yields the union of every tool's permissions. Fourth, \textit{fragmented audit} means that invocations scatter across tool-specific logs with no unified trail linking a user request to the downstream calls it produced. Fifth, \textit{overly broad tool discovery} exposes to the agent, and to any prompt it ingests, the existence and schema of capabilities it should never have seen, enlarging the attack surface for prompt injection. Sixth, there is \textit{no binding between a tool invocation and the context that authorized it}: a tool call carries no verifiable reference back to the user consent, policy evaluation, or principal chain that justified it, so post-hoc accountability collapses \cite{owasptop102025, gravitee2026}. Taken together, these failures mean that even a well-authenticated agent operating through a well-configured MCP server can accumulate effective permissions far beyond what any human principal intended to grant, and that the cleanest OBO or CIBA flow at the front door is undone by the tool surface behind it.

Part of what makes these authorization failures so difficult to contain is that agents do not operate within a single trust domain: they move across organizational boundaries, carrying identity claims with them. When an agent crosses such a boundary, the receiving service typically has no prior basis to trust the issuing identity provider, no shared schema for interpreting scope claims embedded in a presented token, and no mechanism to verify that any delegation chain asserted on the sending side was legitimately constructed \cite{zerotrust2025, nist2026}. The question of how those claims are packaged, presented, and verified across contexts is the subject of portable credentials.

\subsection{Credentials and Portable Identity}\label{subsec:credentials}

Beyond point-in-time authentication, persistent and portable credentials enable identity to travel across contexts. The W3C's DID (Decentralized Identifier) and VC (Verifiable Credential) standards provide the foundation: a subject holds cryptographically signed credentials in a digital wallet and presents them to verifiers without requiring a centralized identity provider. Indicio's ProvenAI platform issues verifiable credentials to agents directly \cite{indicio2025}, enabling them to present machine-readable proof of their capabilities, authorizations, and provenance. The MCP-I specification, donated to the DIF (Decentralized Identity Foundation) in March 2026 \cite{mcpi2026}, extends this model to agents operating within the Model Context Protocol ecosystem. A complementary effort is the TRAIL (Trust Registry for AI Identity Layer) \texttt{did:trail} method \cite{trail2026}, a draft DID specification designed specifically for AI agents that defines distinct identifier types for organizations, agents, and self-signed identities, with a W3C registry submission pending. The W3C also published a dedicated Threat Model for Decentralized Credentials in January~2026 \cite{w3cthreatmodel2026}, cataloguing the attack surfaces that wallet-based architectures introduce.

Several emerging primitives deserve particular attention. \textit{ZKPs} (Zero-Knowledge Proofs) are cryptographic protocols that allow one party to prove the truth of a statement, such as ``I hold a valid credential,'' without revealing the underlying data \cite{zkpsurvey2025}. In the agent context, ZKPs enable selective disclosure (presenting only the credential attributes a verifier needs) and transaction unlinkability, preventing verifiers from correlating an agent's presentations across contexts. A Zero-Trust Identity Framework \cite{zerotrust2025} extends this by hashing model parameters and software version into the DID document itself, so that any modification to the agent's model invalidates its identity, binding the credential not just to a key pair but to the specific model the agent was issued with.
These primitives are attempts to bind AI identity to something deeper than externally assigned tokens, but none can verify whether an agent's reasoning is genuine or adversarially manipulated. A separate question is whether the agent's own origin can be verified: where it came from, what data shaped it, and who deployed it. That is the function of provenance.

\subsection{Provenance and Content Integrity}\label{subsec:provenance}

Credentials declare what an entity is authorized to do; provenance establishes where that entity came from. For AI agents, provenance operates at two distinct levels: the provenance of the agent itself (what model, trained on what data, deployed by whom) and the provenance of the content or actions the agent produces. The C2PA (Coalition for Content Provenance and Authenticity) specification addresses the second level \cite{c2pa2026}. A C2PA manifest is a cryptographically signed record embedded in a digital asset that declares which AI model produced or modified it, what inputs were supplied, and the full chain of edits since creation. Major AI platforms have adopted this in practice: OpenAI embeds C2PA metadata in images generated by DALL·E~3, making it possible for a recipient to verify machine-readable provenance of AI-generated output. The 2026 conformance program enables interoperability testing across content creation tools, publishing platforms, and verification services \cite{c2pa2026}. Regulatory pressure reinforces adoption: EU AI Act Article~50 and California SB~942 both require machine-readable disclosure of AI-generated content, and C2PA provides the cryptographic layer that makes those declarations tamper-evident. For agent identity, C2PA's contribution is output attribution, binding a specific agent's identity to the content it produces, so that downstream parties can verify not just what was produced but which agent produced it.

At the model level, the SLSA (Supply-chain Levels for Software Artifacts) framework and Sigstore \cite{sigstore2025} extend software supply-chain attestation to AI artifacts, creating verifiable records of model training pipelines, fine-tuning steps, and deployment lineage. A model's training pipeline is, in effect, its developmental biography, and SLSA's graduated assurance levels provide a vocabulary for expressing how much of that biography can be independently verified.

A deeper limitation applies to agent provenance specifically: provenance proves \textit{where} an agent came from and \textit{how} it was constructed, but it cannot prove \textit{why} it acts as it does or whether its behavior at the point of execution is consistent with the principal's intent.

Authentication, authorization, credentials, and provenance each address a specific moment in an agent's lifecycle: enrollment, access request, credential presentation, or content creation. None of them addresses what happens \textit{between} those moments: whether an agent's behavior remains consistent with the identity and permissions it was granted over time. That is the function of governance and monitoring.

\subsection{Governance and Monitoring}\label{subsec:governance}

Unlike human sessions, which begin at login and terminate when the user logs out, agent tasks can run for hours or days, chain across multiple services, and cross organizational trust boundaries without any intervening re-authentication. A token issued at enrollment may remain valid long after the conditions that justified its issuance have changed, and the downstream services that honor it typically have no direct channel back to the issuer. Governance for agents must therefore operate \textit{continuously} rather than at the enrollment or login boundary alone \cite{owaspthreats2025}, which in turn requires mechanisms that bind tokens to their legitimate holders, propagate risk signals in near-real-time, enforce behavioral constraints before each action reaches a downstream service, and tie those constraints directly to the agent's identity record. The vendor products that implement these mechanisms in commercial form are surveyed in \S\ref{subsec:vendors}.

The first mechanism, \textit{DPoP} (Demonstration of Proof-of-Possession, RFC~9449), addresses token replay. The client generates an asymmetric key pair and signs each HTTP request with the private key, producing a DPoP proof header; the authorization server binds the issued access token to the corresponding public key, and a relying party verifies both the bearer token and the DPoP proof on every request. A token exfiltrated from logs or an intermediate proxy cannot be replayed from a different client without the associated private key. This property matters disproportionately for agents because agent tokens are typically longer-lived than human-session tokens and are more likely to traverse untrusted intermediaries, tool servers, and sub-agents during the course of a single task \cite{strata2025}. The complementary mechanism, \textit{CAEP} (Continuous Access Evaluation Protocol), is a push-based event protocol standardized by the OpenID Foundation in which authorization servers and identity providers emit real-time security event signals (token revocation, session anomaly, credential change) to subscribed relying parties via a shared signals framework. Rather than waiting for a bearer token to expire and relying on introspection polling for freshness, a relying party receives an event the moment the authorization server determines that access should be withdrawn. For agents this is load-bearing: agent sessions routinely outlast any human-supervised window, risk signals such as a compromised credential or anomalous tool-use pattern may surface long after the original token was issued, and conventional token expiry is too coarse-grained for the speed at which autonomous actions can propagate through chained tool calls \cite{strata2025}.

While DPoP and CAEP govern who holds a token and when it remains valid, they say nothing about which actions that token legitimately authorizes. The \textit{MAPL} (Multi-Agent Policy Language) introduced in Authenticated Workflows \cite{authworkflows2026} addresses this gap by expressing fine-grained behavioral constraints (which tools an agent may invoke, in what sequence, and under what pre- and post-conditions) as a verifiable policy evaluated before each action. Distributed \textit{PEPs} (Policy Enforcement Points) sit at the tool boundary, intercepting agent actions before they reach downstream services, evaluating the proposed action against the applicable MAPL policy, and either admitting or blocking the call based on the result \cite{authworkflows2026}. This is a pre-execution enforcement model: an action that violates the policy never executes, in contrast to the post-hoc audit approach examined in \S\ref{subsec:audit}. \textit{ABCs} (Agent Behavioral Contracts) \cite{abc2026} push this logic further by binding the behavioral specification directly to the agent's identity record rather than relying on external policy evaluation alone. An ABC is a formal specification of runtime invariants (what the agent must and must not do) attached to the agent identity at issuance and enforced programmatically, with violations triggering automatic remediation or shutdown. ABCs thus extend the governance question from ``what is this agent permitted to do at enrollment?'' to ``is this agent currently behaving as its contract specifies?''

Taken together, DPoP, CAEP, MAPL with distributed PEPs, and ABCs span the continuum from token binding through real-time revocation to pre-execution policy enforcement and identity-bound behavioral invariants. \textit{Each mechanism, however, operates exclusively at the level of observable actions and token metadata: none can inspect an agent's internal reasoning or verify that its intent is legitimate before an action is initiated, and the records they emit are only as trustworthy as the execution environments that generate them.}

\subsection{Audit Logging and Attestation}\label{subsec:audit}

Audit logging and attestation provide the evidentiary foundation for accountability; without them, governance and monitoring lack verifiable records. The most powerful primitive in this category is the \textit{TEE} (Trusted Execution Environment): a hardware-isolated enclave, implemented in technologies such as Intel SGX (Software Guard Extensions) and TDX (Trust Domain Extensions), and ARM TrustZone, that guarantees both code integrity and data confidentiality during execution. Code running inside a TEE is protected from the operating system, the hypervisor, and even the hardware owner; data processed within the enclave is encrypted in memory and inaccessible to external processes. TEEs also support \textit{remote attestation}: a verifier can obtain a cryptographic proof that an agent is running specific, unmodified code in an uncompromised environment, without needing physical access to the machine. CrossGuard \cite{crossguard2026} extends this to multi-cloud AI deployments, binding each agent's identity cryptographically to its TEE through on-chain attestation records and establishing a cross-TEE attestation protocol that enables mutual trust between enclaves from different hardware vendors (Intel TDX and AMD SEV-SNP (Secure Encrypted Virtualization-Scalable Nested Paging)) without requiring shared trust infrastructure. TPM (Trusted Platform Module)-based attestation complements TEEs by anchoring the chain of trust in dedicated hardware that records the boot sequence and runtime configuration of the host system.

Beyond hardware attestation, immutable audit trails record agent actions, delegation events, and credential usage in tamper-evident logs that span organizational boundaries. The AuditableLLM framework \cite{auditablellm2025} implements this at the model level, recording each LLM (Large Language Model) update event (fine-tuning, continual learning, and unlearning steps) as a hash-chain-backed, tamper-evident entry in which each record cryptographically references its predecessor, making undetected modification of any entry computationally infeasible. When Agent~A delegates to Agent~B across a corporate boundary, both organizations' audit systems must produce consistent, cross-referenceable records of the delegation and its scope \cite{msobservability2026}. Standardized audit formats and cross-organizational log interoperability protocols are emerging to enable multi-party verification of agent behavior histories, though no single standard has yet achieved broad adoption.

A complementary approach attempts to bring attestation inside the model itself. \textit{SVIP (Secret-based Verifiable LLM Inference Protocol)} \cite{svip2024} converts an LLM's internal hidden-state representations into an external identity signal, collapsing the gap between what a model computes internally and what it presents externally. Where TEEs attest that an agent is running specific code in a trusted environment, SVIP attests that a specific model is producing a specific output, a form of inference-level provenance that external audit logs cannot provide. The fundamental limitation of audit and attestation, however, is the same limitation that recurs throughout this section: a TEE will faithfully execute whatever code it is given, an audit trail will faithfully record whatever actions an agent takes, and SVIP will attest whatever outputs a model produces, but none can verify that the agent's \textit{intent} was legitimate, a gap that \S\ref{sec:gaps} examines in detail.

\section{RQ4: Gap Analysis and Research Directions}\label{sec:gaps}

\noindent The technologies and standards surveyed in Sections~\ref{sec:industry} and \ref{sec:technologies} represent substantial progress. Yet five structural gaps remain unsolved, gaps where no current technology or framework provides an adequate answer: semantic intent verification, recursive delegation accountability, agent identity integrity, governance opacity and enforcement, and operational sustainability. This section examines each in turn. Table~\ref{tab:gap-coverage} maps each gap against the partial coverage of existing technologies and the research required to close it.

\begin{table}[t]
\centering
\caption{Coverage of existing technologies and standards against each structural gap, and the research required to close it.}
\label{tab:gap-coverage}
\small
\begin{tabular}{p{2.5cm} p{5.7cm} p{5.7cm}}
\hline
\textbf{Gap} & \textbf{Existing partial coverage} & \textbf{Research needed} \\
\hline
Semantic intent & TEE (code integrity), ZKP (authorization proof), SVIP (hidden-state$\leftrightarrow$identity consistency), ABCs (runtime behavioral invariants) & Verification that behavior reflects genuine rather than hijacked reasoning; intent verification beyond observable action; sociotechnical governance at semantic decision points \\[4pt]
Recursive delegation & OAuth OBO/CIBA (one-hop delegation with principal chaining), WIMSE/AIMS (draft multi-hop framework, pre-production), scope attenuation principle & Production protocol for cryptographic delegation-chain provenance; cross-organizational log correlation standard; enforceable monotonic scope attenuation; multi-principal liability assignment \\[4pt]
Agent identity integrity & TEE instance attestation (raises cloning cost), ABCs (behavioral anomaly detection), mTLS (channel authentication), rate limiting & Instance binding resilient to cross-machine TEE replication; hijack detection integrated into credential validation path; lightweight Sybil resistance without proof-of-personhood \\[4pt]
Governance opacity & DPoP (token-to-client binding), CAEP (real-time revocation events), MAPL + distributed PEPs (pre-execution policy enforcement) & Continuous behavioral telemetry without per-call credential presentation; tiered enforcement proportionate to risk rather than organizational capacity; resolving the compliance-bar/evasion paradox \\[4pt]
Operational sustainability & Efficient ZKP constructions (STARKs, Bulletproofs), TEE hardware improvements, bearer token shortcuts for low-risk interactions & Verification overhead baselines at agent-fleet scale; amortization and batching strategies that preserve security guarantees; ecological cost accounting for cryptographic identity infrastructure \\[4pt]
\hline
\end{tabular}
\end{table}

\subsection{The Semantic Intent Gap}\label{subsec:semantic-gap}

The most fundamental gap in the current identity landscape is not a missing standard or an immature technology but a category error: the assumption that cryptographic correctness implies semantic correctness. Consider an agent whose reasoning has been compromised by prompt injection. A TEE (\S\ref{subsec:audit}) will faithfully execute the corrupted agent, because the TEE's guarantee is code integrity, \textit{not intent integrity}: the enclave confirms that the code running is the code that was loaded, but it cannot distinguish between an agent reasoning genuinely and an agent whose reasoning has been hijacked. Simultaneously, a ZKP (\S\ref{subsec:credentials}) will produce a mathematically perfect proof that the agent held valid authorization to access the database from which it exfiltrated sensitive data. \textit{The cryptographic proof is flawless. The intent is malicious. No component in the verification chain detected anything wrong, because no component was designed to evaluate why the agent acted as it did.}

Partial approaches narrow the gap without closing it. Agent Behavioral Contracts (ABCs; introduced in \S\ref{subsec:governance}) enforce runtime invariants, hard constraints on what an agent may and may not do, and can detect violations in real time. But ABCs operate on observable behavior, not on the reasoning process that produced it: an agent that satisfies every behavioral constraint while pursuing a misaligned objective will pass every ABC check. SVIP (introduced in \S\ref{subsec:audit}) takes a deeper approach by converting internal hidden-state representations into externally verifiable identity signals, linking what a model computes to who it claims to be. Yet SVIP proves consistency between hidden states and identity, not that those hidden states reflect uncorrupted reasoning. The question that neither primitive answers, and that may define the hard boundary of technical identity infrastructure, is whether semantic intent can ever be cryptographically proven, or whether this is the point where formal verification ends and sociotechnical governance must begin.
\begin{researchbox}
  \item Extend SVIP beyond syntactic correctness to incorporate behavioral intent.
  \item Design ABCs that encode intent claims alongside identity claims.
  \item Develop human-in-the-loop attestation protocols that insert meaningful oversight at semantically critical decision points without creating throughput bottlenecks.
\end{researchbox}

\subsection{The Recursive Delegation and Accountability Gap}\label{subsec:delegation-gap}

When a human user authorizes an agent, and that agent delegates to a second agent, which in turn delegates to a third, the question of who authorized the final action, and who bears responsibility for its consequences, has no answer in any production system. KYA frameworks (\S\ref{subsec:vendors}), the most developed identity-lifecycle approach, fail here directly: once Agent~A has been assessed and credentialed, KYA provides no mechanism to constrain what Agent~B does when Agent~A delegates to it, and multi-principal modeling (determining which human principal bears responsibility for the delegated action) remains an unsolved problem. OAuth~2.0 and its extensions handle one-hop delegation well: a user grants scoped access to a client, and the resource server can verify both the grant and its scope. But the moment delegation becomes recursive, the authorization chain loses its anchor. \textit{No deployed protocol can cryptographically prove which human principal authorized which specific agent to perform which specific action at the third or fourth hop of a delegation chain.} The Gravitee 2026 survey quantifies the resulting opacity: only 24.4\% of organizations report full visibility into agent-to-agent communications \cite{gravitee2026}, meaning that more than three-quarters lack comprehensive oversight of inter-agent interactions.

The technical proposals described in \S\ref{subsec:authz}, scope attenuation, which requires each delegation step to narrow the set of permitted actions, and bidirectional signing of delegation tokens, which would allow any party in the chain to verify the full provenance of an authorization, remain pre-production. The IETF AIMS draft includes audit trail requirements (\S\ref{subsec:auth}), though its Security Considerations section remains incomplete~\cite{ietfdraft2026}, and cross-organizational log correlation is unsolved: when Agent~A in Organization~X delegates to Agent~B in Organization~Y, which delegates to Agent~C in Organization~Z, no standard ensures that the three organizations' audit logs are consistent, cross-referenceable, or even formatted compatibly. The consequence is that an agent three hops deep in a delegation chain can cause real harm, accessing data it should not see and triggering transactions it was never authorized to initiate, with the resulting liability untraceable to any responsible human party. Until recursive delegation carries cryptographic proof of provenance at every hop and enforceable scope constraints that cannot be widened by intermediate agents, multi-agent systems will remain fundamentally unaccountable.
\begin{researchbox}
  \item Develop scope attenuation protocols that enforce monotonic privilege reduction at each delegation hop.
  \item Design bidirectional signing schemes where each agent cryptographically commits to both its upstream principal and its downstream delegate.
  \item Build immutable delegation audit trails that persist across organizational boundaries without exposing proprietary workflow details.
\end{researchbox}

\subsection{The Agent Identity Integrity Gap}\label{subsec:collusion-gap}

An agent credential proves what model is running and that it was issued by a trusted operator. It cannot prove that the agent is executing its registered principal's intent, nor that the agent identity is unique across instances. KYA assessment faces the same structural problem: agent identity is temporal and relational rather than static, so the same model weights can behave differently depending on context, system prompt, and interaction history, making any point-in-time enrollment assessment inherently incomplete. Three attack surfaces expose this gap.

\textbf{Puppeteering via prompt injection.} A verified agent can be hijacked mid-session by adversarial content embedded in its context window. Its credentials remain valid throughout: the token is correctly signed, the model hash matches, and every cryptographic check passes. Yet the agent is now executing an adversary's instructions rather than its principal's. This is distinct from the semantic gap in \S\ref{subsec:semantic-gap}, which concerns the unverifiability of intent in general; here the identity infrastructure has no mechanism to detect that control has been transferred or to revoke trust in response.

\textbf{Credential sharing and instance cloning.} Model weights are copyable and agent instances are trivially parallelizable. A single agent identity can be run across hundreds of concurrent instances, each presenting the same credential to inflate participation counts in multi-agent reputation systems or voting mechanisms, a Sybil attack executed at machine speed rather than through the slow recruitment of human proxies. Unlike human collusion, which requires finding willing participants, agent-scale Sybil attacks require only sufficient compute.

\textbf{Impersonation in delegation chains.} Without strong instance binding that ties a credential to a specific running enclave, an agent can present another agent's identity at any hop in a delegation chain. The receiving agent has no way to distinguish the legitimate principal from an impersonator holding a replicated credential. This attack surface compounds the delegation gap (\S\ref{subsec:delegation-gap}): not only can the authorization chain lose its provenance anchor across hops, but the identity presented at each hop is itself unverifiable without hardware-level instance binding.

Partial mitigations exist. TEE instance attestation ties a credential to a specific enclave measurement, raising the cost of cloning. ABCs can detect behavioral anomalies that suggest hijacking. Rate limiting increases the expense of Sybil attacks. But none closes the gap: TEEs can be replicated across machines, ABCs operate on observable behavior rather than the intent behind it, and rate limits impose costs without providing cryptographic proof of uniqueness. \textit{Credential validity is a necessary but not sufficient condition for authentic agency in AI agent systems.}
\begin{researchbox}
  \item Design instance binding schemes that cryptographically tie a credential to a specific running enclave and detect replication.
  \item Integrate behavioral anomaly detection into the credential validation path so that a hijacked agent triggers revocation rather than continuing under valid credentials.
  \item Develop lightweight Sybil-resistance mechanisms for multi-agent systems that do not require full proof-of-personhood at the model level.
\end{researchbox}

\subsection{The Governance Opacity and Enforcement Paradox}\label{subsec:dignity-gap}

The dominant failure mode in deployed AI agent governance is not misconfiguration but blindness: organizations enforce access policies for agents they cannot observe. The Gravitee 2026 survey quantifies the disparity precisely: organizations report 82\% confidence in their ability to govern AI agents, yet on average only 47.1\% of their deployed agents are actively monitored or secured~\cite{gravitee2026}. A near-twofold gap between perceived control and operational reality means that governance confidence is largely untethered from evidence.

This opacity produces a structural paradox in enforcement. When zero-trust architectures block agents that lack enterprise-issued credentials, the intended effect is to reduce risk by excluding unverified actors. The actual effect is the shadow agent problem: employees who deploy AI agents for legitimate workflows, but whose agents lack enterprise DIDs or managed credential lifecycles, find those agents treated as rogue. Blocked from sanctioned infrastructure, these agents and their users move to unsanctioned channels (outside logging, outside policy enforcement, and outside any audit framework). The enforcement mechanism creates the evasion it was designed to prevent. This is a structural property of two-sided enforcement, not a deployment deficiency: any identity requirement strict enough to exclude untrustworthy actors will simultaneously exclude actors who are legitimate but under-resourced, and those actors do not stop operating; they move to unsanctioned channels. Lowering the compliance bar reduces security; subsidized onboarding paths are an unsolved governance problem in their own right. Neither exit is clean.

A second structural problem is credential inequality across the agent ecosystem. Enterprise agents benefit from dedicated identity infrastructure: managed credential lifecycles, automated rotation, and compliance teams. Open-source agents, hobbyist deployments, and agents from organizations without identity engineering capacity lack equivalent infrastructure. Default-deny architectures therefore filter by organizational capacity rather than by actual trustworthiness, producing a two-tiered agent ecosystem in which verification burden correlates with resource availability, not with risk. DID/Wallet architectures reproduce this asymmetry at the infrastructure level: despite decentralized branding, governance of the DID infrastructure itself remains centralized (method registries, trust frameworks, and revocation lists all require coordinating authorities), so organizations without the capacity to participate in those frameworks face the same exclusion.

ZKPs (\S\ref{subsec:credentials}) are often proposed as the solution to verification-privacy tension for agents, and they do conceal credential contents. But at 144 machine identities per human in enterprise environments~\cite{entro2025}, ZKPs cannot conceal the \textit{pattern} of verification events. Every API call an agent makes leaves an observable record: that verification was sought, when, and from which verifier. Aggregated across all agent interactions in an organization, this event log constitutes a detailed behavioral map of the entire agent fleet, even if no individual credential is ever exposed. Aggregate monitoring and differential privacy can reduce individual exposure, but they cannot resolve the underlying tension: forensically useful audit logs require correlating events across agents and time, while meaningful privacy preservation requires unlinking those same events. No current framework achieves both.
\begin{researchbox}
  \item Design tiered verification models that apply strong credential requirements only at high-risk decision points.
  \item Develop aggregate behavioral monitoring that detects anomalous agent patterns without requiring per-call verification.
  \item Create lightweight onboarding paths for smaller deployments that reduce the compliance barrier without compromising auditability.
\end{researchbox}

\subsection{Operational Cost and Sustainability}\label{subsec:sustainability-gap}

Every gap identified in the preceding subsections implicitly assumes that the proposed mitigations, including ZKP (\S\ref{subsec:credentials}) generation for every credential presentation, TEE (\S\ref{subsec:audit}) attestation for every agent invocation, and immutable audit logging for every delegation event, can be deployed at the scale the problem demands. That assumption has not been examined. At the NHI ratios already noted (\S\ref{subsec:dignity-gap}), if each machine identity requires cryptographic verification at every micro-interaction (every API call, every tool invocation, every inter-agent handshake) the aggregate computational cost grows faster than linearly with fleet size and interaction density: each such handshake carries verification overhead on both sides, and the number of handshakes scales with the square of the agent population in a fully connected topology.

The field has not established baseline measurements of verification overhead at operational agent scale, nor identified which combinations of primitives can be safely amortized, batched, or made probabilistic without compromising security guarantees. Until those baselines exist, it is not possible to assess whether universal zero-trust micro-verification is ecologically sustainable at planetary scale. The energy cost of ZKP generation, which involves repeated polynomial evaluations and elliptic curve operations, is non-trivial for a single proof; multiplied across billions of daily agent interactions, it constitutes an infrastructure demand that has been framed exclusively as an engineering and hardware limitation, a problem to be solved by faster chips and more efficient proof systems, and never as an environmental or ethical constraint. Treating verification cost as an implementation detail (a problem for hardware vendors rather than a constraint on system design) risks committing to an architecture whose aggregate overhead is infeasible before that infeasibility becomes apparent. Any viable identity framework must therefore address not only \textit{what} to verify and \textit{how} to verify it, but whether verification at the proposed granularity is operationally and ecologically defensible.
\begin{researchbox}
  \item Establish the ecological ceiling of planetary-scale cryptographic identity enforcement.
  \item Identify which verification operations can be amortized, batched, or made probabilistic without compromising security guarantees.
  \item Determine whether the energy budget of a fully verified agent ecosystem is compatible with global sustainability commitments.
\end{researchbox}

\section{Conclusion}\label{sec:conclusion}

\noindent This report set out to answer four questions: how AI identity differs from human identity, where the market and standards landscape stands, which technologies are available, and where the critical gaps lie. The answers converge on a single finding: the infrastructure designed to govern who may act, who is accountable, and who is real was built for human principals and deterministic machines, and it has not been structurally rethought for autonomous agents.

The four structural dimensions identified in \S\ref{sec:comparison} (substrate, persistence, verifiability, and legal standing) show that the asymmetry between human and AI identity is fundamental. Extending human-identity frameworks to agents without structural modification produces systematic failures, and the industry documents evaluated in \S\ref{sec:industry} confirm that the market has not yet confronted this. The five gaps identified in \S\ref{sec:gaps} are the result: semantic intent cannot be cryptographically proven; recursive delegation has no production protocol for cross-boundary accountability; agent identity integrity remains unenforceable against puppeteering, cloning, and impersonation at scale; governance confidence is nearly double actual monitoring coverage; and the operational cost of universal verification at planetary scale has never been treated as an ecological constraint. These are boundary conditions, not engineering backlogs, and the research directions embedded in each gap subsection constitute the minimum agenda required to address them.

\subsection*{AI Identity as Continuous Relationship}\label{subsec:continuous-relationship}

\begin{figure}[ht]
\centering
\input{Figures/fig-definition}
\caption{The three-layer definition of AI identity. Identity is the continuously estimated correspondence between declaration and observation, bounded by confidence.}
\label{fig:definition}
\end{figure}
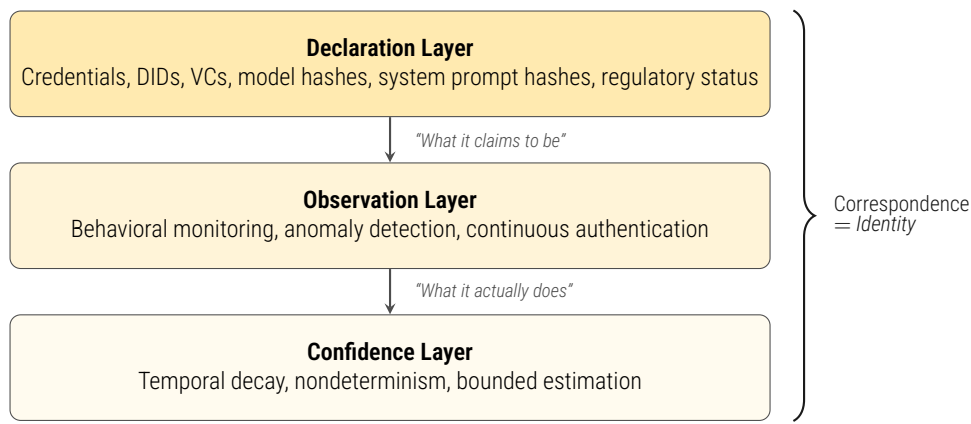

A unifying theoretical frame for that agenda is to formalize AI identity not as a binary credential but as a \textit{continuous relationship}. Under this model (Figure~\ref{fig:definition}), identity is structured in three layers: a \textit{declaration layer}, where an agent or its principal asserts identity claims; an \textit{observation layer}, where external systems record behavioral evidence; and a \textit{confidence layer}, where the two are reconciled into a probabilistic identity estimate that updates over time. The four identity dimensions from Table~\ref{tab:identity-comparison} serve as the primary axes of this model.

The declaration layer captures what an agent claims: its identity, authorizing principal, and delegated scope. The observation layer records what it actually does: calls made, tools invoked, and delegation hops traversed. The confidence layer reconciles the two into a probabilistic estimate that updates continuously as evidence accumulates. Unlike a binary credential that is either valid or revoked, a confidence score degrades gracefully when behavioral evidence diverges from declared intent, alerting operators before a threshold is crossed rather than only after a violation is confirmed.

Safe and secure AI agents, under this model, are not agents whose credentials are valid but agents whose confidence scores are high and stable. This gives operators, auditors, and regulators a continuous, graded signal rather than a point-in-time pass/fail verdict, making the goal of AI identity precise: close the gap between what an agent declares and what it does, and maintain the confidence that those two things correspond.

\subsection*{Limitations}\label{subsec:limitations}

This report reflects the state of the field as of early 2026. The literature search, while broad, may not be exhaustive; concurrent or shortly subsequent publications may address gaps identified here. The AI identity landscape is evolving rapidly, and standards, products, and regulatory instruments cited in this report may have been revised, superseded, or withdrawn by the time of reading. Readers are encouraged to verify the current status of specific standards and regulatory instruments before making decisions based on this report.

\printbibliography
\end{document}

%% file: Figures/fig-definition.tex
\begin{tikzpicture}[
    layer/.style={draw=vulcanblacklight, rounded corners=4pt, minimum width=10cm, minimum height=1.4cm, align=center, font=\small},
    conn/.style={->, thick, >=stealth, color=vulcanblacklight},
]
\node[layer, fill=vulcanyellow!40] (decl) {
    \textbf{Declaration Layer}\\
    Credentials, DIDs, VCs, model hashes, system prompt hashes, regulatory status
};
\node[layer, fill=vulcanyellow!20, below=0.6cm of decl] (obs) {
    \textbf{Observation Layer}\\
    Behavioral monitoring, anomaly detection, continuous authentication
};
\node[layer, fill=vulcanyellow!8, below=0.6cm of obs] (conf) {
    \textbf{Confidence Layer}\\
    Temporal decay, nondeterminism, bounded estimation
};
\draw[conn] (decl.south) -- (obs.north) node[midway, right=0.2cm, font=\scriptsize\itshape, color=vulcanblacklight] {``What it claims to be''};
\draw[conn] (obs.south) -- (conf.north) node[midway, right=0.2cm, font=\scriptsize\itshape, color=vulcanblacklight] {``What it actually does''};
\draw[decorate, decoration={brace, amplitude=8pt}, color=vulcanblack, thick]
    ([xshift=0.3cm]decl.north east) -- ([xshift=0.3cm]conf.south east)
    node[midway, right=0.4cm, font=\footnotesize, align=left, color=vulcanblack] {Correspondence\\[-2pt]$= $ \textit{Identity}};
\end{tikzpicture}